\documentclass{article}
\usepackage{spconf,amsmath,graphicx}
\usepackage{booktabs}
\usepackage{multirow}
\usepackage{amssymb}
\usepackage{pifont}
\usepackage{hyperref}

\usepackage{enumitem}
\setlist{nosep, leftmargin=14pt}

\usepackage{mwe} 

\title{FreqDINO: Frequency-Guided Adaptation for Generalized Boundary-Aware Ultrasound Image Segmentation}
\name{
    \begin{tabular}{@{}c@{}}
    Yixuan Zhang$^{1,*}$, Qing Xu$^{1,2,*}$, Yue Li$^{1,2,*}$, Xiangjian He$^{1,\dag}$, Qian Zhang$^{1,\dag}$, Mainul Haque$^{1}$ \\
    Rong Qu$^{2}$, Wenting Duan$^{3}$, Zhen Chen$^{4}$\thanks{$^*$Equal contribution. $^\dag$Corresponding author.}
    \end{tabular}
}
\address{$^1$University of Nottingham Ningbo China, $^2$University of Nottingham, \\ $^3$ University of Lincoln, $^4$Yale University}
\begin{document}
%
\maketitle
\begin{abstract}
Ultrasound image segmentation is pivotal for clinical diagnosis, yet challenged by speckle noise and imaging artifacts. Recently, DINOv3 has shown remarkable promise in medical image segmentation with its powerful representation capabilities. However, DINOv3, pre-trained on natural images, lacks sensitivity to ultrasound-specific boundary degradation. To address this limitation, we propose FreqDINO, a frequency-guided segmentation framework that enhances boundary perception and structural consistency. Specifically, we devise a Multi-scale Frequency Extraction and Alignment (MFEA) strategy to separate low-frequency structures and multi-scale high-frequency boundary details, and align them via learnable attention. We also introduce a Frequency-Guided Boundary Refinement (FGBR) module that extracts boundary prototypes from high-frequency components and refines spatial features. Furthermore, we design a Multi-task Boundary-Guided Decoder (MBGD) to ensure spatial coherence between boundary and semantic predictions. Extensive experiments demonstrate that FreqDINO surpasses state-of-the-art methods with superior achieves remarkable generalization capability. The code is at \url{https://github.com/MingLang-FD/FreqDINO}.

\end{abstract}

\section{Introduction}
\label{sec:intro}

Ultrasound image segmentation plays a crucial role in clinical applications such as breast cancer detection and thyroid nodule diagnosis, where accurate boundary delineation directly impacts diagnostic reliability and treatment planning precision. However, ultrasound imaging is inherently challenged by speckle noise, low signal-to-noise ratio, and acoustic shadowing artifacts that result in blurred and discontinuous boundaries \cite{chen2022aau, zhang2024low}, making precise segmentation extremely challenging. Therefore, developing robust segmentation methods capable of accurately capturing fine boundary details under such degradations is therefore a critical need in clinical practice.

Early ultrasound segmentation methods primarily relied on convolutional neural networks (CNNs) such as U-Net \cite{ronneberger2015u} and its variants \cite{isensee2021nnu, chen2022aau, rahman2024emcad, chen2024ma} to capture anatomical structures through multi-scale features. Subsequently, transformer-based approaches \cite{chen2024transunet} achieved significant progress by modeling long-range dependencies through self-attention mechanisms. The recent advent of vision foundation models has further revolutionized medical image analysis, with models like SAM series \cite{kirillov2023segment, ravi2025sam} demonstrating remarkable zero-shot capabilities and DINOv3 \cite{simeoni2025dinov3} exhibiting powerful self-supervised representation learning on natural images. These foundation models have shown great potential in medical imaging tasks, offering opportunities for improved generalization. Unlike SAM, which relies on manual interactive prompts, DINOv3 offers a fully convolution-free vision transformer trained through self-distillation, producing dense, high-quality features ideal for fine-grained segmentation adaptation.

\begin{figure*}[!t]
  \centering
  \includegraphics[width=0.9\linewidth]{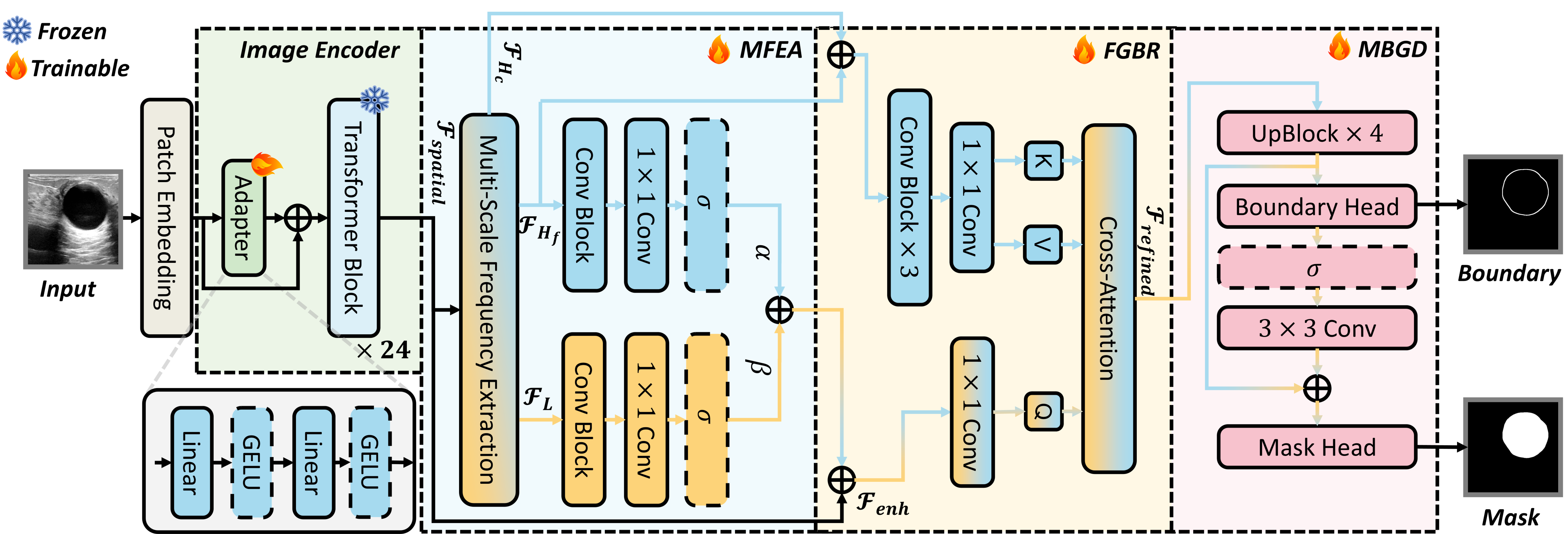}
  \caption{The overview of our FreqDINO framework for ultrasound image segmentation, consisting of MFEA, FGBR, and MBGD. FreqDINO adapts DINOv3 by explicitly leveraging frequency-domain decomposition for precise boundary perception.}
  \label{fig:method}
\end{figure*}

\begin{figure}[!t]
	\begin{center}
		\includegraphics[width=0.8\linewidth]{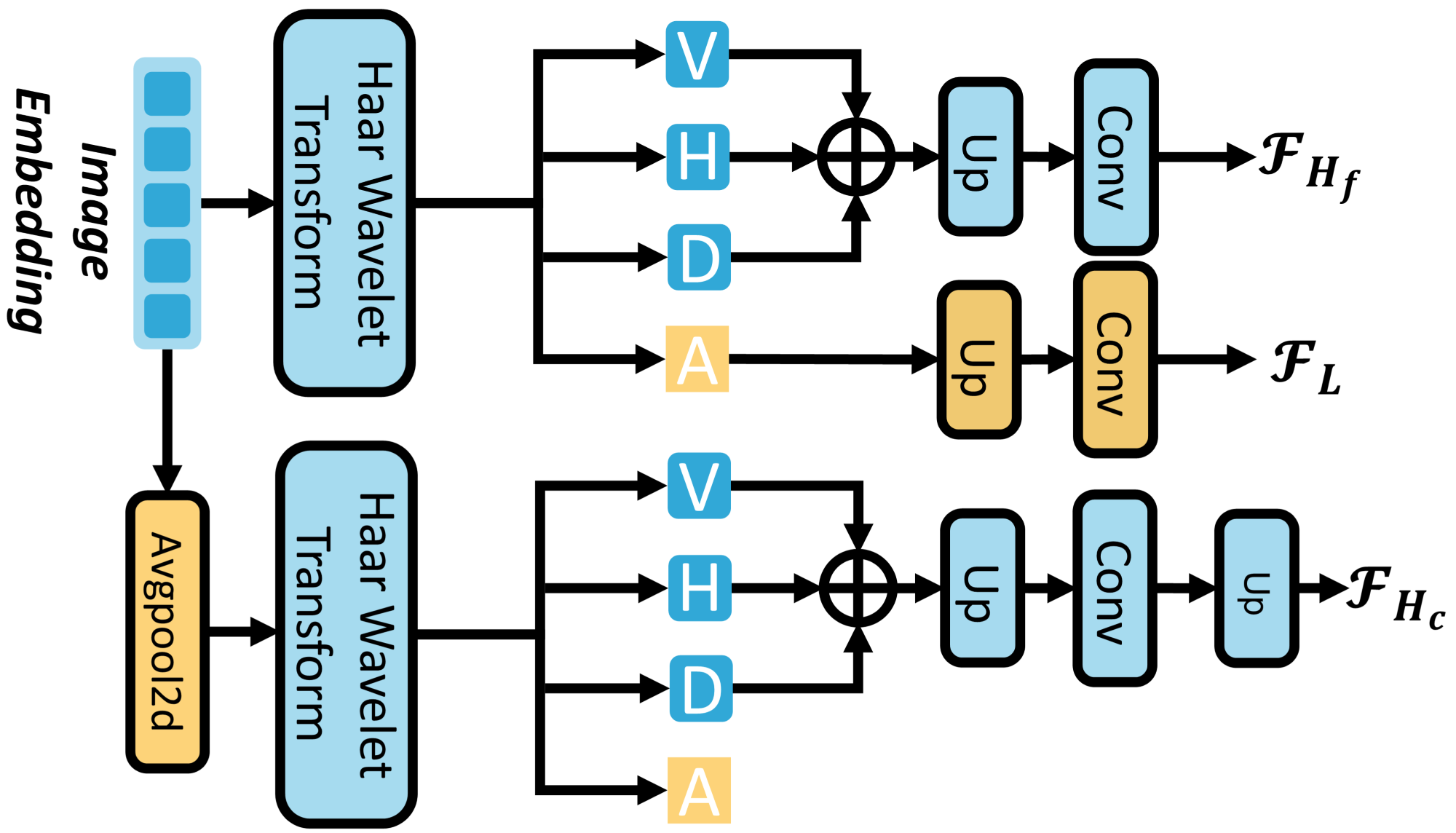}
	\end{center}
	\caption{The detailed illustration of multi-scale frequency extraction in the MFEA of our FreqDINO framework.}
	\label{fig:method2}
\end{figure}

Despite the progress, DINOv3 still lacks the perception of blurred boundaries in ultrasound. This limitation stems from the fundamental difference between natural images and medical ultrasound: ultrasound boundaries are naturally encoded in the frequency domain, with high-frequency components corresponding to sharp boundary transitions and low-frequency components representing smooth anatomical structures \cite{meyer2025ultrasam}. However, DINOv3 operates purely in the spatial domain where boundary and structure cues are implicitly entangled, limiting its ability to perceive ultrasound-specific frequency patterns. More critically, the domain gap between natural image pre-training and ultrasound data, such as speckle noise patterns and low-contrast transitions, further hinders effective boundary perception. Explicitly leveraging frequency-domain decomposition thus offers a promising avenue for adapting DINOv3 to ultrasound segmentation.

To address these limitations, we propose FreqDINO, a frequency-guided segmentation framework that adapts DINOv3 for ultrasound imaging through explicit frequency decomposition and boundary enhancement. Specifically, a Multi-scale Frequency Extraction and Alignment (MFEA) module disentangles low-frequency structures and multi-scale high-frequency boundaries via Haar wavelet transform \cite{wu2025wavelet}, and fuse them via learnable boundary-structure attention. A Frequency-Guided Boundary Refinement (FGBR) module extracts boundary prototypes from high-frequency components and refines spatial features through cross-modal attention for precise boundary guidance. Furthermore, a Multi-task Boundary-Guided Decoder (MBGD) with a dual-head architecture jointly optimizes boundaries and semantic predictions through multi-task supervision. In this way, these synergistic modules collectively enable FreqDINO to achieve precise and generalizable boundary delineation across diverse ultrasound imaging conditions. Extensive experiments demonstrate that FreqDINO outperforms state-of-the-art methods with superior boundary localization and remarkable zero-shot generalization capability.

\section{Methodology}
\label{sec:format}

\subsection{Overview of FreqDINO}
As shown in Fig.\ref{fig:method}, FreqDINO adapts DINOv3 for ultrasound segmentation through frequency-guided boundary enhancement. Given an input ultrasound image, we first extract spatial features using a frozen DINOv3 encoder with lightweight adapters for parameter-efficient transfer. The proposed framework then introduces three synergistic components: 1) MFEA decomposes spatial features into high-frequency boundaries and low-frequency structures via Haar wavelet transform at different scales, producing enhanced features through learnable boundary-structure attention. 2) FGBR extracts boundary prototypes from high-frequency components and refines features via cross-attention. 3) MBGD employs dual-head architecture with multi-task supervision for joint boundary and semantic prediction. By integrating these modules, FreqDINO effectively compensates for DINOv3’s spatial-domain limitation, enabling precise boundary perception and robust generalization across diverse ultrasound imaging.

\subsection{Multi-Scale Frequency Extraction and Alignment}
Ultrasound images exhibit distinct frequency characteristics where low-frequency components encode anatomical structures while high-frequency components capture boundary details. Direct spatial feature learning struggles to distinguish these complementary patterns. Given spatial features $\mathcal{F}_{\text{spatial}} \in \mathbb{R}^{B \times C \times H_1 \times W_1}$ from DINOv3, we employ haar wavelet decomposition at two scales, At the original $H_1 \times W_1$ resolution, we decompose features into low-frequency structure $\mathcal{F}_{LL}$ and three high-frequency components $\{\mathcal{F}_{LH}, \mathcal{F}_{HL}, \mathcal{F}_{HH}\}$ (horizontal, vertical, and diagonal) encoding boundary details:
\begin{gather}
    \mathcal{F}_{H_f} = \phi_H(\text{Concat}[\mathcal{F}_{LH}, \mathcal{F}_{HL}, \mathcal{F}_{HH}]), \\ 
    \mathcal{F}_L = \phi_L(\mathcal{F}_{LL}),
\end{gather}
where $\phi_H$ and $\phi_L$ are $1\times1$ convolutions for channel reduction. To capture multi-scale patterns, we extract coarse-grained boundary features $\mathcal{F}_{H_c}$ at $H_2\times W_2$ resolution through downsampling and upsampling. We then generate boundary attention $\mathcal{A}_b$ from $\mathcal{F}_{H_f}$ and structure attention $\mathcal{A}_s$ from $\mathcal{F}_L$ via lightweight networks, and combine them with learnable weights $\alpha = \beta = 0.5$. The enhanced features are computed through residual modulation with fusion weight $\lambda=0.3$:
\begin{equation}
\mathcal{F}_{\text{enh}} = \mathcal{F}_{\text{spatial}} + \lambda \cdot (\mathcal{F}_{\text{spatial}} \odot (\alpha \mathcal{A}_b + \beta \mathcal{A}_s)).
\end{equation}

\subsection{Frequency-Guided Boundary Refinement.} 
While MFEA captures multi-scale frequency patterns, explicit boundary knowledge transfer remains challenging due to high-dimensional feature complexity. We address this through boundary prototype distillation. We distill a $64$-dimensional boundary prototype from concatenated high-frequency features $\mathcal{F}_{H_f}$ and $ \mathcal{F}_{H_c}$ through progressive dimensionality reduction. The cross-modal attention mechanism queries this prototype using enhanced spatial features, where query $\mathbf{Q}$ comes from $\mathcal{F}_{\text{enh}}$ and key-value pairs come from boundary prototype. Using $8$-head attention with per-head dimension $128$, the refined features are obtained via residual fusion with learnable weight $\omega=0.2$:
\begin{equation}
    \mathcal{F}_{\text{refined}} = \mathcal{F}_{\text{enh}} + \omega \cdot \mathbf{W}_O(\text{Attn}(\mathbf{Q}, \mathbf{K}, \mathbf{V})).
\end{equation}
This two-stage design ensures both global frequency awareness and precise boundary guidance while preserving DINOv3's semantic richness.

\begin{table}[!t]
\centering
\setlength\tabcolsep{3pt}
\caption{Comparison with state-of-the-arts on BUSI.}
\label{tab:internal results}
\begin{tabular}{l|ccc}
\hline
Methods & Dice (\%) $\uparrow$ & mIoU (\%) $\uparrow$  & HD (mm) $\downarrow$ \\
\hline
UNet \cite{ronneberger2015u} \ & 71.22 & 59.58 & 155.55 \\
UNext \cite{valanarasu2022unext} & 78.32 & 68.53 & 82.62 \\
nnU-Net \cite{isensee2021nnu} & 84.80 & 76.44 & 46.63 \\
AAU-Net \cite{chen2022aau} & 81.32 & 71.67 & 55.57 \\
TransUNet \cite{chen2024transunet} & 75.28 & 64.96 & 88.57 \\
EMCAD \cite{rahman2024emcad}  & 75.13 & 64.98 & 63.58  \\
MADGNet \cite{nam2024modality} & 80.09 & 70.03 & 61.49 \\
\hline
SAM \cite{kirillov2023segment}  & 78.42  & 69.76  & 84.81  \\
SAM2 \cite{ravi2025sam}  & 78.52  & 68.56  & 72.29  \\
Med-SA \cite{wu2023medical} & 82.60  & 74.62  & 63.26  \\
SAM2-Adapter \cite{chen2025sam2} & 81.64  & 73.20  & 68.22  \\
MedSAM \cite{ma2024segment}  & 70.91  & 60.79  & 107.16  \\
UltraSam \cite{meyer2025ultrasam} & 75.82  & 66.48  & 94.88  \\
\hline
FreqDINO & \textbf{86.52} & \textbf{78.49} & \textbf{39.63} \\
\hline
\end{tabular}
\end{table}

\subsection{Multi-Task Boundary-Guided Decoder}
To ensure spatial consistency between semantic and boundary predictions, we adopt a dual-head decoder that jointly learns mask and boundary representations. Given $\mathcal{F}_{\text{refined}} \in \mathbb{R}^{B \times C \times H_1 \times W_1}$ from FGBR, we progressively upsample through four $2 \times 2$ transposed convolution blocks $\mathcal{F}_{\rm shared}=\text{UpBlock}(\mathcal{F}_{\text{refined}})$. We employ a boundary-first strategy: the boundary prediction $\mathcal{M}_{\text{boundary}} = \text{Conv}_{1\times1}(\mathcal{F}_{\text{shared}})$ is first generated, then converted to boundary features $\mathcal{F}_{\text{boundary}} = \text{Conv}_{3\times3}(\sigma(\mathcal{M}_{\text{boundary}}))$, where $\sigma$ denotes sigmoid. Finally, mask prediction leverages both features via concatenation:
\begin{equation}
\mathcal{M}_{\text{mask}} = \text{Conv}_{1\times1}(\mathcal{F}_{\text{shared}} \oplus \mathcal{F}_{\text{boundary}}).
\end{equation}
This boundary-guided design ensures accurate segmentation with well-defined boundaries.

\subsection{Optimization Pipeline}
Our training follows a multi-task learning paradigm that jointly optimizes mask and boundary predictions. The framework employs a frozen DINOv3 encoder with lightweight adapters for parameter-efficient adaptation from natural images to the ultrasound domain, while the frequency modules (MFEA and FGBR) and decoder MBGD are trained end-to-end. Since pixel-level boundary annotations are unavailable, we automatically generate boundary ground truth from mask annotations using morphological operations:

The training objective combines mask segmentation and boundary prediction through a weighted multi-task loss:
\begin{equation}
    \mathcal{L}_{\text{total}} = \mathcal{L}_{\text{mask}} + \lambda_b \cdot \mathcal{L}_{\text{boundary}},
\end{equation}
where $\mathcal{L}_{\text{mask}}$ and $\mathcal{L}_{\text{boundary}}$ are binary cross-entropy losses, and $\lambda_b = 0.3$. By jointly optimizing $\mathcal{L}_{\text{total}}$, FreqDINO achieves accurate ultrasound segmentation with precise boundaries.

\begin{table}[!t]
\centering
\setlength\tabcolsep{3pt}
\caption{Generalization on the unseen TN3K dataset.}
\label{tab:external results}
\begin{tabular}{l|ccc}
\hline
Methods & Dice (\%) $\uparrow$ & mIoU (\%) $\uparrow$ & HD (mm) $\downarrow$ \\
\hline
UNet \cite{ronneberger2015u} & 35.38 & 24.85 & 188.50 \\
UNext \cite{valanarasu2022unext} & 41.56 & 31.93 & 153.47 \\
nnU-Net \cite{isensee2021nnu} & 54.94 & 45.33 & 120.33 \\
AAU-Net \cite{chen2022aau}& 41.73 & 32.36 & 142.91 \\
TransUNet \cite{chen2024transunet} & 45.36 & 34.50 & 146.66 \\
EMCAD \cite{rahman2024emcad}  & 42.17 & 31.96 & 135.77 \\
MADGNet \cite{nam2024modality} & 43.28 & 33.35 & 145.23  \\
\hline
SAM \cite{kirillov2023segment}  & 55.70  & 45.13  & 129.99  \\
SAM2 \cite{ravi2025sam}  & 56.55  & 45.73  & 126.79  \\
Med-SA \cite{wu2023medical}  & 60.70  & 50.78  & 114.03  \\
SAM2-Adapter \cite{chen2025sam2}  & 54.28  & 44.50  &  126.73 \\
MedSAM \cite{ma2024segment}  &  52.56 & 41.67  & 133.25  \\
UltraSam \cite{meyer2025ultrasam}  & 60.70  & 45.96  &  139.39 \\
\hline
FreqDINO & \textbf{62.09} & \textbf{51.94} & \textbf{108.01} \\
\hline
\end{tabular}
\end{table}

\section{Experiments}
\label{sec:format}

\subsection{Experimental Setup}
We evaluate our framework on two public ultrasound datasets: BUSI \cite{al2020dataset} and TN3K \cite{gong2023thyroid}. BUSI is a breast ultrasound segmentation dataset containing $780$ images from $600$ female patients at an average resolution of $500\times500$. Since only the benign and malignant cases include segmentation masks, we use the annotated $647$ images from benign and malignant categories for internal validation, split them into training, validation, and test sets with a $8:1:1$ ratio. TN3K is a thyroid nodule ultrasound segmentation dataset comprising $3,493$ images from $2,421$ patients captured with various devices, with resolutions ranging from $216\times217$ to $1463\times771$ pixels, serving as external validation to assess generalization capability. All images are resized to $512\times512$ for unified processing. All experiments are conducted on an NVIDIA A5000 GPU using PyTorch. Our model employs the DINOv3-Large encoder, while comparison methods use their respective large-scale variants to ensure fair comparison. We use the Adam optimizer with an initial learning rate of $\times 10^{-4}$ and exponential decay (factor $0.98$). Training is performed with a batch size $16$ for $300$ epochs. For evaluation, we adopt three standard ultrasound segmentation metrics: Dice coefficient for segmentation overlap, mean Intersection over Union (mIoU), and Hausdorff Distance (HD).

\subsection{Comparison with State-of-the-Art Methods}
\begin{figure}[!t]
  \centering
  \includegraphics[width=1\linewidth]{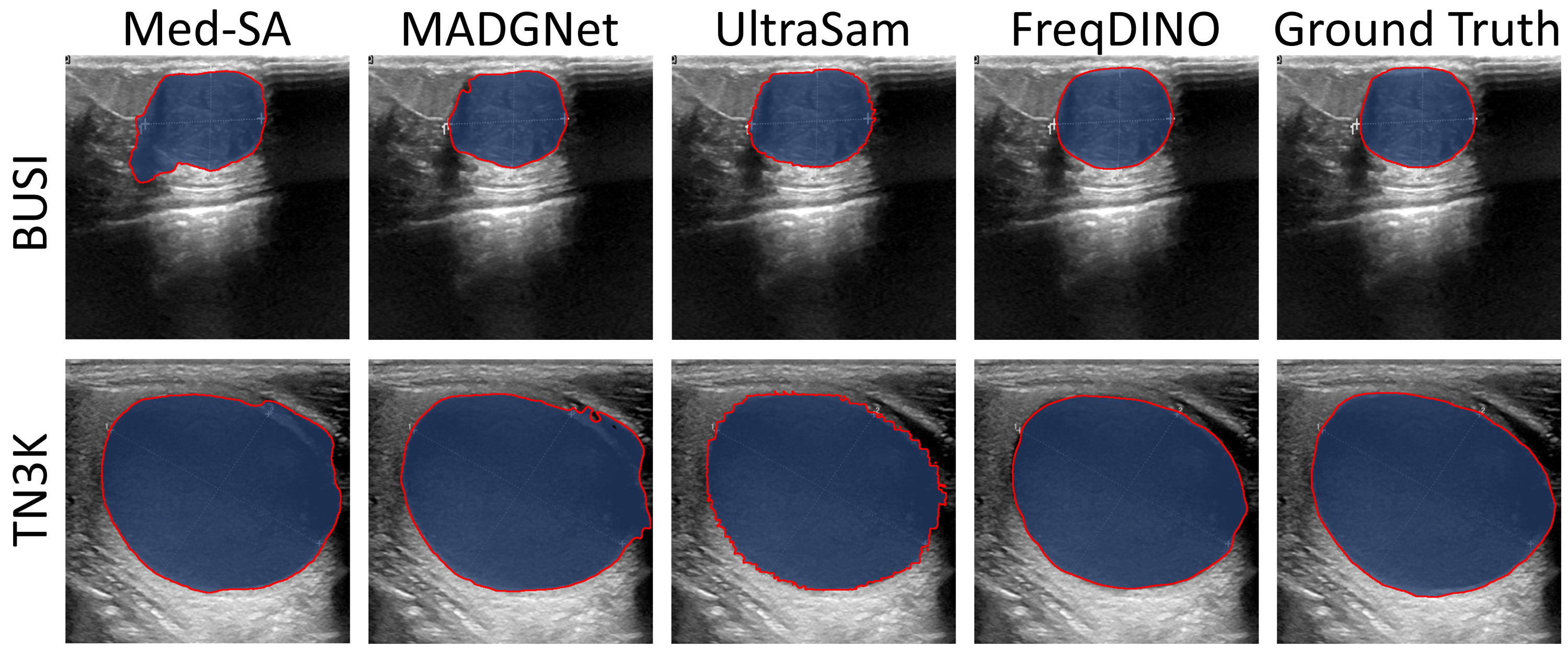}
  \caption{Visualization comparison of ultrasound image segmentation on BUSI and TN3K datasets. Our FreqDINO exhibits the best results, achieving more accurate boundary localization with precise edge delineation while suppressing speckle noise interference and reducing false positives.}
  \label{fig:visual}
\end{figure}

We conduct comprehensive comparisons on the BUSI dataset against classical segmentation methods and foundation model-based approaches. For fair comparison, all fully fine-tuned U-Net series models and SAM-based methods adopt the no-prompt inference setting. As illustrated in Table \ref{tab:internal results}, classical segmentation methods achieve comparable performance to foundation model-based approaches. Remarkably, nnU-Net demonstrates strong performance, surpassing Med-SA with a $2.66\%$ Dice increase and $16.63mm$ HD reduction, indicating the effectiveness of specialized medical image segmentation architectures. FreqDINO achieves the best performance across all metrics, with a Dice score of $86.52\%$ and the lowest HD of $39.63mm$. Compared to the second-best nnU-Net, FreqDINO further improves Dice by $2.01\%$ and reduces HD by $7.00mm$, highlighting the benefit of frequency-guided boundary modeling. Qualitative comparisons in Fig.~\ref{fig:visual} further show that the proposed FreqDINO can delineate boundaries more accurately with better edge precision.

\subsection{Zero-Shot Generalization Analysis}

We further evaluate generalization capability through zero-shot inference on the TN3K dataset without fine-tuning. As shown in Table \ref{tab:external results}, foundation models significantly outperform classical methods, contrasting with the comparable performance in Table \ref{tab:internal results}. Notably, nnU-Net shows limited generalization, underperforming Med-SA ($60.70\%$ Dice, $114.03mm$ HD) by $10.48\%$ Dice despite its strong in-domain performance. Our FreqDINO demonstrates the strongest generalization with $62.09\%$ Dice and $108.01mm$ HD, further improving upon Med-SA by $2.29\%$ Dice and $6.02mm$ HD reduction. The substantial HD improvement particularly highlights the effectiveness of explicit frequency-domain boundary guidance in maintaining precise edge delineation across different ultrasound imaging protocols.

\subsection{Ablation Study}
\begin{table}[!t]
\centering
\setlength\tabcolsep{2pt}
\caption{Ablation study of FreqDINO on the BUSI dataset.}
\label{tab:ablation study results}
\begin{tabular}{l|ccc}
\hline
MFEA FGBR MBGD & Dice (\%) $\uparrow$ & mIoU (\%) $\uparrow$ & HD (mm) $\downarrow$ \\
\hline
& 82.35 & 72.39 & 47.59\\
\hspace{10pt}\checkmark & 84.17 & 74.62 & 44.59 \\
\hspace{10pt}\checkmark \hspace{20pt} \checkmark & 85.13 & 76.76 & 43.02 \\
\hspace{10pt}\checkmark \hspace{20pt} \checkmark \hspace{20pt} \checkmark  & \textbf{86.52} & \textbf{78.49}& \textbf{39.63}\\
\hline
\end{tabular}
\end{table}
To evaluate the contribution of each component in FreqDINO, we conduct an ablation study on BUSI using DINOv3-Large with adapters as a baseline. As shown in Table \ref{tab:ablation study results}, introducing MFEA achieves $2.21\%$ Dice improvement and $3.00mm$ HD reduction. Combined MFEA and FGBR further improve performance ($1.14\%$ Dice, $1.57mm$ HD reduction), demonstrating the effectiveness of frequency decomposition and boundary-guided refinement work complementarily to enhance boundary perception. The complete FreqDINO with MBGD achieves $86.52\%$ Dice and $39.63mm$ HD, showing that our frequency-domain guidance framework effectively enhances boundary-aware segmentation. These results validate that the tailored MFEA, FGBR, and MBGD collectively contribute to the superior performance of FreqDINO.

\section{conclusion}
\label{sec:format}
In this work, we proposed FreqDINO, a frequency-guided framework that adapts DINOv3 for ultrasound image segmentation by explicitly leveraging frequency-domain information to enhance boundary perception. The model integrates three complementary modules: MFEA for extracting multi-scale high-frequency boundaries and aligning frequency components to enhance spatial features, FGBR for refining features via boundary prototypes distilled from high-frequency components, and MBGD for ensuring spatial consistency through boundary-guided mask generation. Extensive experiments on BUSI and TN3K datasets demonstrate that FreqDINO outperforms state-of-the-art methods with superior boundary localization and achieves strong generalization capability.

\bibliographystyle{IEEEbib}
\bibliography{refs}

\end{document}